\documentclass{article}
\usepackage{graphicx}
\usepackage{amsmath} 
\usepackage{amssymb} 
\usepackage{amsfonts}
\usepackage{algorithmic}
\usepackage{graphicx} 
\usepackage{color}
\usepackage{booktabs}
\usepackage{tabularx}
\usepackage{wrapfig}
\usepackage{textcomp}
\usepackage{csvsimple}
\usepackage{hyperref}
\usepackage{balance}
\usepackage{nomencl}
\usepackage{float}
\usepackage{subcaption}
\makenomenclature
\usepackage[table]{xcolor}
\usepackage[parfill]{parskip}
\usepackage{fancyhdr}
\usepackage{authblk}
\usepackage[backend=biber]{biblatex}
\usepackage{multirow}
\usepackage{microtype}

\begin{document}

\title{Sex-based Network-Specific Differences in Connectomes: A Krakencoder-Based Analysis}

\author[1]{Vibhashree S H\thanks{vibha.hul@gmail.com}}
\author[2]{Debanjali Bhattacharya\thanks{b\_debanjali@blr.amrita.edu}}
\author[1]{Vamshi Krishna Kancharla\thanks{vamshi.kancharla461@gmail.com}}
\author[1]{Neelam Sinha\thanks{neelam@cbr-iisc.ac.in}}

\affil[1]{Centre for Brain Research, Indian Institute of Science, Bengaluru 560012, India}
\affil[2]{Dept. of Artificial Intelligence, Amrita School of Artificial Intelligence, Amrita Vishwa Vidyapeetham, Bengaluru 560035, India}

\date{15 June 2026}

\maketitle

\begin{abstract}
    This study examines how deficiencies in one brain connectome modality propagate to the other, using the Krakencoder as a simulation framework. Structural and functional connectomes from 702 healthy participants in the Human Connectome Project were analyzed, with the impact of each of the Yeo-7 functional networks assessed separately. Seven scenarios were considered, each involving the removal of a single network while the remaining networks were preserved. The resulting perturbations in cross-modal predictions were quantified using three complementary metrics: KL divergence on eigenvalue spectra, Frobenius norm, and Wasserstein distance. In addition, the persistence of sex-specific information within the predicted connectomes was evaluated. Across all metrics and both prediction directions, the Default Mode Network produced the largest perturbations, whereas the Somatomotor network yielded the smallest. Sex differences in network-level perturbation signatures were subtle, with the best result being an accuracy of 66.09\% from connectomes predicted under network-removal conditions. In contrast, connectomes predicted from intact inputs achieved substantially higher sex classification accuracy, reaching up to 84.76\%. These findings confirm that full predicted connectomes retain considerably more sex-discriminative information than perturbation-derived signatures alone.
\end{abstract}

\section*{Keywords}
Structural connectome, Functional connectome, Krakencoder, Sex classification, Network analysis, Network ablation analysis, KL divergence, Frobenius norm, Wasserstein distance

\section{Introduction}

\subsection{Connectome Basics}
The human brain can be partitioned \cite{baldassarre2012individual} \cite{eickhoff2018imaging} into multiple regions of interest (ROIs), which may be directly or indirectly connected to one another. A direct connection indicates the presence of white matter tracts structurally linking two regions. When these tracts are reconstructed from diffusion MRI (dMRI) data using tractography methods, and the strength of each connection is quantified, the resulting representation is termed a structural connectome (SC). In parallel, the extent of indirect or statistical associations between regions can be measured, yielding a functional connectome (FC).

%\subsection{Mathematical Properties of Connectomes}
Both SCs and FCs are represented as $n \times n$ matrices, where $n$ denotes the number of ROIs in the chosen parcellation scheme. Each element of an SC matrix encodes a tractography-derived weight, such as streamline count or streamline density, and all entries are non-negative, since negative structural connections are not physically meaningful. In contrast, each element of an FC matrix corresponds to the normalised covariance between a pair of regions: positive values reflect correlated activity, while negative values reflect anticorrelated activity. SC matrices are symmetric, sparse, and non-negative, whereas FC matrices are symmetric, dense, and positive semidefinite. The properties of these matrices are summarised in Table ~\ref{tab:scfcprops}.

%\subsection{Utilities of SCs and FCs}
SCs have been widely applied to the study of white matter development, ageing, and structural degeneration in neurological conditions. FCs are sensitive to cognitive state, psychiatric disorders, and demographic variables such as age and sex, and have been extensively used in brain–behaviour mapping and clinical classification. Both modalities encode individual-specific information that generalises across sessions, making them reliable neuroimaging fingerprints.

\begin{table}[h]
\centering
\caption{Comparison of structural and functional connectome properties.}\label{tab:scfcprops}
\setlength{\tabcolsep}{12pt}
\makebox[\textwidth][c]{%
\begin{tabular}{lll}
\toprule
Property & SC & FC \\
\midrule
Elements represent & Streamline count/weight & Normalised covariance \\
Value range & Non-negative and zero only & Positive, zero, and negative \\
Matrix properties & Symmetric, sparse & Symmetric, dense, PSD \\
Clinical sensitivity & WM development, ageing & Cognition, psychiatry, sex \\
\bottomrule
\end{tabular}
}
\end{table}

\subsection{Krakencoder}
The Krakencoder is a tool introduced by Jamison et al. \cite{jamison2025krakencoder}, to predict and consolidate connectome matrices across modalities. Different parcellation schemes, preprocessing pipelines, and connectivity quantification methods provide distinct perspectives on the same underlying connectivity data, referred to as “flavours.” The Krakencoder is capable of handling 15 primary flavours, of which two are utilised in this study (see Section~\ref{sec:supplementary}). The model consists of encoders and decoders spanning the 15 aforementioned flavours. By mapping each flavour into a shared latent hypersphere and decoding back into any other flavour, Krakencoder achieved substantially higher individual-level identifiability than prior SC--FC models, with improvements of 42--54\%. The latent space preserved demographic and behavioral information such as age, sex, and cognition, and generalized without retraining to out-of-distribution datasets. This work established Krakencoder as a versatile architecture for multimodal connectome fusion.

\subsection{SC–FC Coupling}
Structurally connected regions generally exhibit stronger functional coupling. However, functional connectivity is not limited to direct anatomical links: regions indirectly connected via shared neighbours may co-activate, and even regions without direct structural connections can display strong functional correlations due to neuromodulatory dynamics and temporal fluctuations in neural state. Sensory regions typically show tight correspondence between SC and FC, whereas higher-order association regions are considerably more variable. Previous models often treated SC-to-FC prediction as a one-way mapping and performed poorly, not because the coupling itself was weak, but because the modelling approaches were limited. The Krakencoder addresses this by treating SC and FC as two incomplete views of the same underlying brain, learning a shared latent space in which both modalities are jointly represented. This fused representation is more informative about cognition and demographics than either modality alone.

In practice, a structural connectome is input to the Krakencoder. Because connectomes are symmetric matrices, the upper triangular elements are extracted and flattened into a single-row vector. After extracting the most salient connectivity patterns, this vector is passed into an encoder that reduces dimensionality and stores the data in a common latent space. A decoder then reconstructs the data in the desired flavour specified by the user.

Huang et al. \cite{huang2026progression} applied Krakencoder to study sex differences across the lifespan. Using 1,286 individuals aged 8--100, they fine-tuned Krakencoder to enhance sex-specific encoding and analyzed classification accuracy across age bins. Their results showed that sex differences are minimal in childhood, peak in young to mid-adulthood, and diverge across modalities in later life. Functional differences were driven by higher-order association networks such as the default mode and control networks, while structural differences were concentrated in cerebellar and subcortical pathways. This study provided a lifespan-wide multimodal map of sex differences in brain networks.

\begin{figure}[H]
    \centering
    \includegraphics[width=1.3\linewidth]{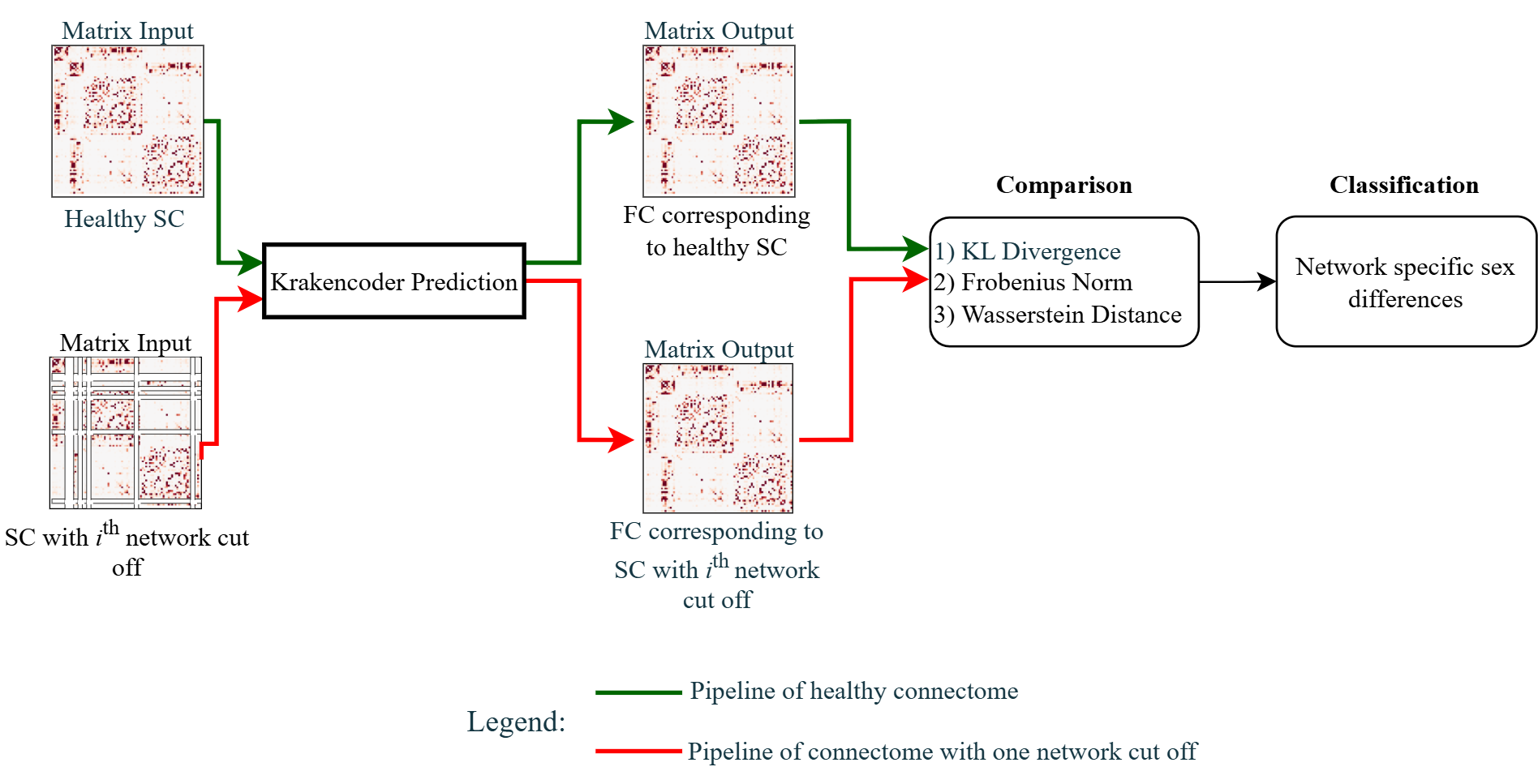}
    \caption{Block diagram of the proposed workflow. For each of the seven Yeo functional networks, all regions belonging to the target network zeroed out in the input connectome to simulate that particular network being cut off.}
    \label{fig:workflow}
\end{figure}

This paper introduces a similar network-level framework applied to the Krakencoder’s learned SC–FC mapping. The framework enables systematic analysis of how the removal of each Yeo-7 functional networks \cite{yeo2011cortex} from the input connectome perturbs the predicted output, thereby characterising the relative contribution of each network to cross-modal connectome translation in both SC-to-FC and FC-to-SC directions. Finally, sex-based differences are investigated using the resulting perturbation signatures, examining whether the Krakencoder’s learned mapping encodes sex-specific structure in a network-resolved manner, similar to the work done by Huang et al. As a baseline measure, sex classification is also performed using the unperturbed structural and functional connectomes. The workflow followed in this study is as shown in Figure~\ref{fig:workflow}.

\section{Dataset}

The dataset used in this study was obtained from the Human Connectome Project (HCP) \cite{hcp_data}, comprising 702 healthy participants (351 male and 351 female). Data acquisition was performed on a Siemens 3T Connectome Skyra scanner using a gradient-echo EPI sequence during resting state with eyes open, while subjects fixated on a bright crosshair. For this study, preprocessed connectome matrices were directly sourced from Kothapalli et al. \cite{kothapalli2025datadriven}

Connectomes were constructed using the Desikan–Killiany atlas \cite{desikan2006automated} \cite{fischl2002whole} with an 86-region parcellation scheme (FS86). Regions were grouped into functional networks according to the Yeo-7 definitions. Data was available for 82 of the 86 regions across both modalities.

Two connectome flavours were utilized in this study. For functional connectivity, partial correlation following global signal regression was used, quantifying direct statistical dependencies between region pairs after removing shared variance attributable to the global signal and all other regions. For structural connectivity, probabilistic tractography with volume normalisation was applied, generating streamlines based on fibre orientation distributions and normalising connection strength by the volume of connected region pairs to correct for region size.

\section{Related Work}

Other studies have focused on sex classification using connectomes. Kothapalli et al. \cite{kothapalli2025datadriven} proposed a multimodal framework combining CNN-based feature extraction with Riemannian tangent space embeddings. Using diffusion-weighted and fMRI-derived connectomes from 787 HCP subjects, they achieved peak accuracy of 93\% when fusing structural and functional features, with Grad-CAM interpretability revealing distinct connectivity patterns differentiating males and females. Zhang et al. \cite{zhang2020gender} combined gray matter volume, regional homogeneity, and functional connectivity, achieving 96.6\% classification accuracy and showing complementary encoding of gender differences across modalities. Cai et al. \cite{cai2025sexual} extended this to white matter functional connectomes, finding females had stronger WM--WM and GM--WM connectivity, with machine learning classifiers effectively distinguishing sex. Dipentima et al. \cite{dipentima2026sex} examined dynamic edge functional connectivity, reporting that males exhibited longer trough durations while females showed higher peak heights, especially in the default mode network. Together, these studies demonstrate that sex differences are robustly encoded in both structural and functional connectomes, though the specific modalities and networks contributing to classification vary.

Network-level contributions to sex differences have also been examined. In their lifespan study, Huang et al. \cite{huang2026progression} performed network inclusion sensitivity analyses, showing that the default mode network consistently carried the strongest sex signal, with cerebellar and subcortical networks becoming increasingly important with age. Similarly, Gu et al. \cite{gu2021bcn} identified the default mode network as most relevant for Alzheimer’s classification using a supervised graph convolutional network, reinforcing its centrality in both demographic and clinical contexts. Beyond sex classification, other works have explored SC--FC prediction and related methodological advances. Goñi et al. \cite{goni2014resting} introduced analytic path-based measures such as search information and path transitivity, achieving correlations up to 0.6. Li et al. \cite{li2022learning} trained a graph convolutional encoder--decoder to jointly reconstruct FC from SC and classify subjects, reaching $\sim$66\% accuracy in distinguishing heavy drinkers from controls. Etemadyrad et al. \cite{etemadyrad2022functional} proposed SF-GAN, integrating subject-level metafeatures and outperforming baselines on Pearson correlation and MSE, with age emerging as the most influential predictor. Tan et al. \cite{tan2026sfc} developed SFC-GAN, a CycleGAN-based bidirectional translator, achieving FC$\rightarrow$SC correlations of 0.94 and supporting downstream classification. Zalesky et al. \cite{zalesky2024predicting} cautioned that individual-level predictions remain modest, with effect sizes around 8--11\% above group averages, emphasizing the need for rigorous benchmarking. Other related studies include Rasero et al. \cite{rasero2018predicting}, who trained classifiers on task-based fMRI to predict intrinsic connectivity networks with $\sim$90\% accuracy when tested on resting-state data, and Suárez et al., who reviewed SC--FC coupling models and concluded that higher-order interactions are needed to explain imperfect correlations. Finally, Ren et al. \cite{ren2013constructing} proposed modified sparse representation for functional connectivity graphs, revealing clearer cognitive-state differences than correlation methods.

\section{Proposed Methodology}

This study investigates sex-based differences in brain network organisation through a network cutting-off approach applied to structural and functional connectomes using the Krakencoder tool. Each of the seven Yeo functional networks is independently removed from the input connectome, one at a time. The Krakencoder then predicts the corresponding output connectome from the degraded input, and the deviation relative to a baseline prediction (obtained from intact connectomes) is quantified using three metrics \cite{bassett2017network}.

Four experimental cases are defined:

\begin{itemize}
    \item \textbf{Case I:} Healthy SC $\rightarrow$ Krakencoder $\rightarrow$ Predicted FC
    \item \textbf{Case II:} Healthy FC $\rightarrow$ Krakencoder $\rightarrow$ Predicted SC
    \item \textbf{Case III:} FC with Yeo network $i$ cut off $\rightarrow$ Krakencoder $\rightarrow$ Predicted SC
    \item \textbf{Case IV:} SC with Yeo network $i$ cut off $\rightarrow$ Krakencoder $\rightarrow$ Predicted FC
\end{itemize}

Here, $i \in \{1, \ldots, 7\}$ indexes the Yeo-7 functional networks: Visual, Somatomotor, Dorsal Attention, Ventral Attention, Limbic, Control (Frontoparietal), and Default Mode.

Cases I and II establish the healthy baseline predictions. In these scenarios, the full, unmodified SC or FC is provided as input to the Krakencoder, and the corresponding predicted output connectome is saved for each subject. These baselines serve as reference points against which perturbations introduced in cases III and IV can be quantified.

Cases III and IV introduce controlled network-level disruptions. For each of the seven Yeo networks, all rows and columns corresponding to that network’s constituent regions are zeroed out in the input connectome, thereby simulating the removal of its contribution in the entire brain. The modified input is then passed through the Krakencoder, and the difference between the resulting prediction and the healthy baseline is computed to yield a difference matrix $\Delta_i$. This matrix captures the deviation induced by the absence of a specific network.

To quantify these deviations, three complementary metrics are computed for each subject and each network: 

\begin{itemize}
    \item KL divergence on eigenvalue spectra, reflecting changes in the variance structure of connectivity assuming a linear relationship between structural and functional connectomes \cite{cover2006elements};
    \item Frobenius norm, measuring total edge-level reconstruction error \cite{janati2020advances};
    \item Wasserstein distance, the latter two metrics assessing nonlinear distributional shifts \cite{muskulus2011wasserstein}.
\end{itemize}

Together, these metrics produce a per-subject, per-network feature vector that characterises the perturbation signature. This constitutes the first contribution of the study: a systematic, network-resolved analysis of how each Yeo-7 network contributes to the Krakencoder’s learned SC–FC coupling.

The second contribution concerns sex-based differences. Both the unmodified predictions from Cases I and II and the perturbation-derived feature sets from Cases III and IV are used independently and in combination for binary sex classification. This enables investigation of whether sex-specific differences in SC–FC coupling are detectable at the network level through the Krakencoder’s latent mapping.

\subsection{Introduction of Network Deficiency}

For each subject, modified versions of both SC and FC matrices are generated by zeroing all rows and columns corresponding to the FS86 regions belonging to a particular Yeo network, applied symmetrically. Diagonal elements are excluded for both modalities. This procedure yields seven modified matrices per modality per subject. A baseline prediction is also computed for each subject using an input with all network connections intact. Two prediction directions are evaluated: 

\begin{itemize}
    \item FC (network cut off) $\rightarrow$ Predicted SC
    \item SC (network cut off) $\rightarrow$ Predicted FC 
\end{itemize}

The difference matrix for network $i$ is defined as:
\begin{equation*}
\Delta_i = A_{\text{cutoff.prediction},i} - A_{\text{baseline.prediction}}
\end{equation*}

\subsection{Performance Metrics}

Three metrics are computed on each difference matrix $\Delta_i$: KL divergence, Frobenius norm, and Wasserstein distance. Each metric captures a distinct aspect of perturbation, allowing evaluation of whether the SC–FC relationship is better described by linear or nonlinear assumptions.

\textbf{KL Divergence on Eigenvalue Spectra}  
KL divergence is used  to quantify the difference between two probability distributions. In this context, the normalised eigenvalue spectra of the network-deficient and baseline predictions are treated as distributions, and their divergence is computed as:
\begin{equation*}
D_{KL}(p \| q) = \sum_k p_k \log \frac{p_k}{q_k}
\end{equation*}
A large KL divergence indicates that cutting a network off substantially redistributes variance across principal connectivity modes. This metric assumes linearity in the SC–FC relationship, as eigenvalues represent variance explained by principal modes. Thus, KL divergence captures whether removing a network alters the variance structure of the predicted connectome under a linear model.

\textbf{Frobenius Norm}  
The Frobenius norm is a matrix norm commonly used to measure reconstruction error in linear algebra and machine learning. It quantifies the total magnitude of edge-level change across the entire difference matrix:
\begin{equation*}
\|\Delta_i\|_F = \sqrt{\sum_{j,k} \Delta_i(j,k)^2}
\end{equation*}
A large Frobenius norm indicates widespread numerical changes across many edges. This metric does not rely on spectral assumptions and directly measures perturbations at the edge level. It is therefore well-suited to capturing nonlinear deviations in SC–FC prediction, reflecting the overall scale of disruption without imposing structural constraints.

\textbf{Wasserstein Distance}  
The Wasserstein distance (or Earth Mover’s Distance) is widely used in optimal transport theory to measure the geometric cost of transforming one distribution into another. Here, it is applied to the distribution of upper-triangle values of $\Delta_i$, comparing them to a distribution with no perturbation:
\begin{equation*}
W_1(u,v) = \inf_{\gamma \in \Gamma(u,v)} \int |x - y| \, d\gamma(x,y)
\end{equation*}
This metric is sensitive to both the shape and location of the perturbation distribution. It reveals whether cutting a network off induces systematic shifts in predicted connectivity values, rather than random noise. By operating directly on the distribution of edge weights, Wasserstein distance captures nonlinear and distributional changes in SC–FC coupling that eigenvalue-based analysis may overlook.

Together, KL divergence reflects linear variance redistribution, while Frobenius norm and Wasserstein distance capture nonlinear, edge-level and distributional perturbations. This combination enables a comprehensive evaluation of how network deficiencies propagate across modalities in Krakencoder predictions.

\subsection{Feature Construction and Classification}

For each subject, a feature vector is constructed by concatenating the three perturbation metrics (KL divergence, Frobenius norm, and Wasserstein distance) computed for each of the seven Yeo networks, yielding a 21-dimensional vector:
\begin{equation*}
\mathbf{f} = \left[ D_{KL}^{(1)},\ \|\Delta_1\|_F,\ W_1^{(1)},\ \ldots,\ D_{KL}^{(7)},\ \|\Delta_7\|_F,\ W_1^{(7)} \right] \in \mathbb{R}^{21}
\end{equation*}

This vector is constructed separately for each prediction direction: $\mathbf{f}_{\text{SC}}$ for the FC (network cut off) $\rightarrow$ Predicted SC direction, and $\mathbf{f}_{\text{FC}}$ for the SC (network cut off) $\rightarrow$ Predicted FC direction. For the combined condition, the two vectors are concatenated, yielding 42 features:
\begin{equation*}
\mathbf{f}_{\text{(SC+FC)}} = [\mathbf{f}_{\text{SC}},\ \mathbf{f}_{\text{FC}}] \in \mathbb{R}^{42}
\end{equation*}

Each of these three feature sets is evaluated under two conditions: networks cut off and networks intact, resulting in six feature sets in total. All feature vectors are Z-score normalised prior to classification. Ten classifiers are evaluated under stratified 5-fold cross-validation: Logistic Regression, SVM (RBF) \cite{cortes1995support}, Random Forest \cite{breiman2001random}, KNN, AdaBoost \cite{freund1997decision}, Gradient Boosting \cite{friedman2001greedy}, LDA \cite{fisher1936use}, Naive Bayes, MLP, and XGBoost \cite{chen2016xgboost}.

\section{Results}

\subsection{Network-Level Perturbation Analysis}

\begin{figure}[H]
\centering
\includegraphics[width=1\linewidth]{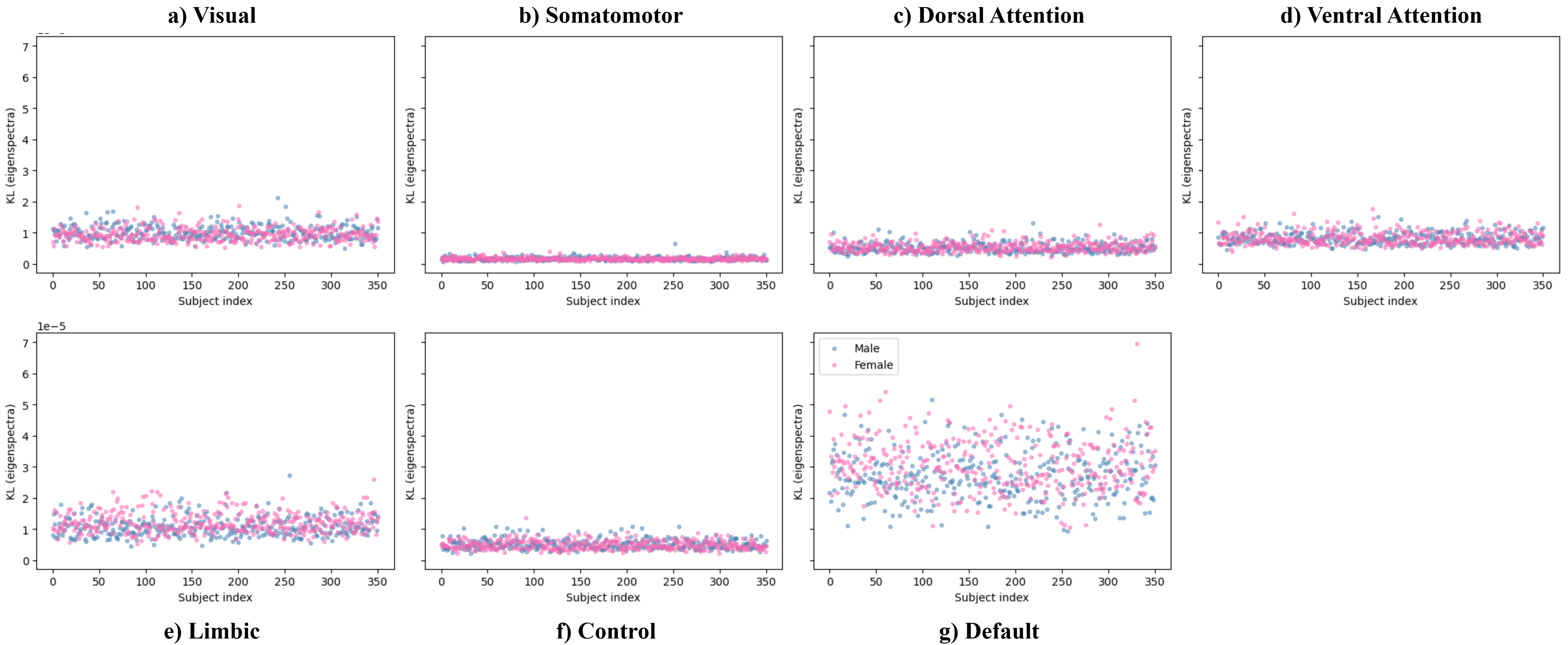}
\caption{KL divergence between eigenvalue spectra of predictions obtained from inputs with networks cut off, and baselines: SC (network cut off) to predicted FC. X-Axis: Subject Index; Y-Axis: KL Divergence (0--7e-5). Blue: male, pink: female. Each subfigure shows the per-subject distribution for one Yeo network.}
\label{fig:klsc2fc}
\end{figure}

\begin{figure}[H]
\centering
\includegraphics[width=1\linewidth]{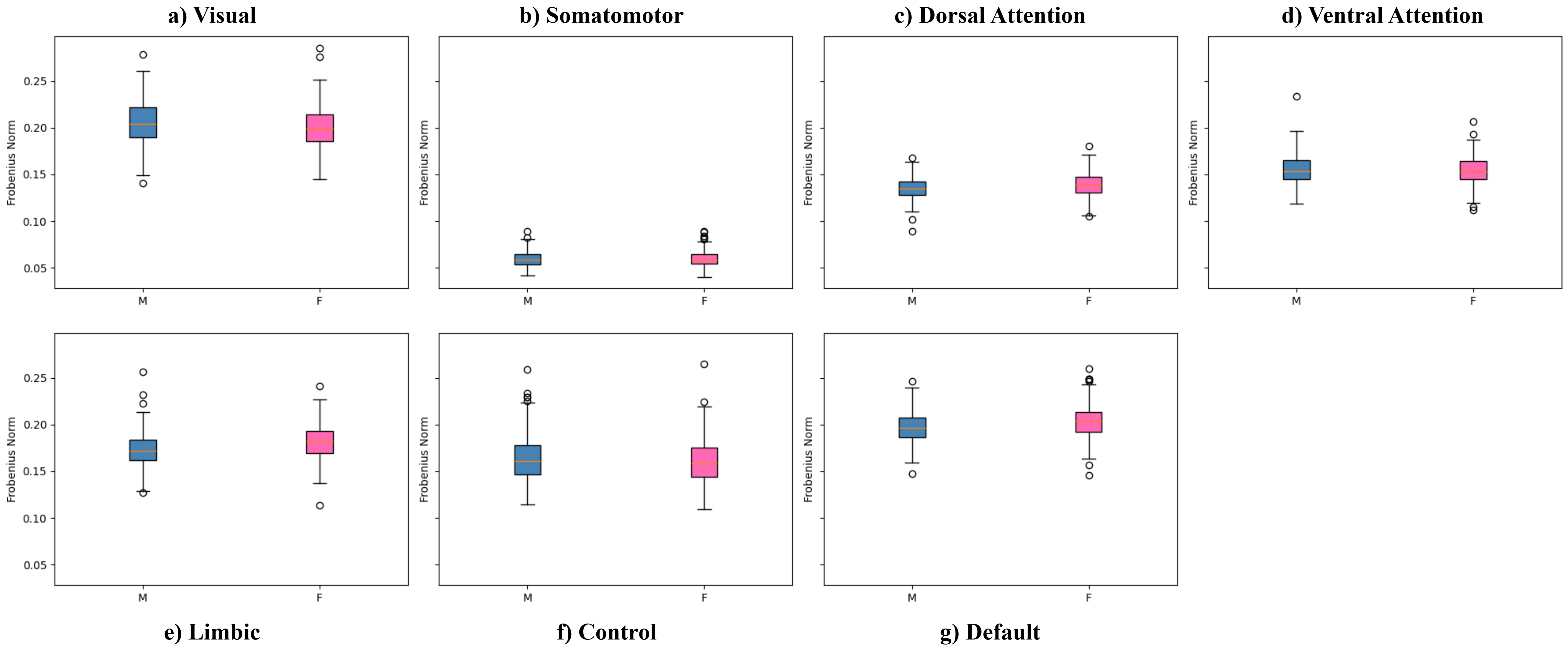}
\caption{Frobenius norm of the difference matrix: SC (network cut off) to predicted FC. X-Axis: Genders; Y-Axis: Frobenius Norm (0.05--0.25). Blue: male, pink: female. Each subfigure shows the per-subject distribution for one Yeo network.}
\label{fig:frobsc2fc}
\end{figure}

\begin{figure}[H]
\centering
\includegraphics[width=1\linewidth]{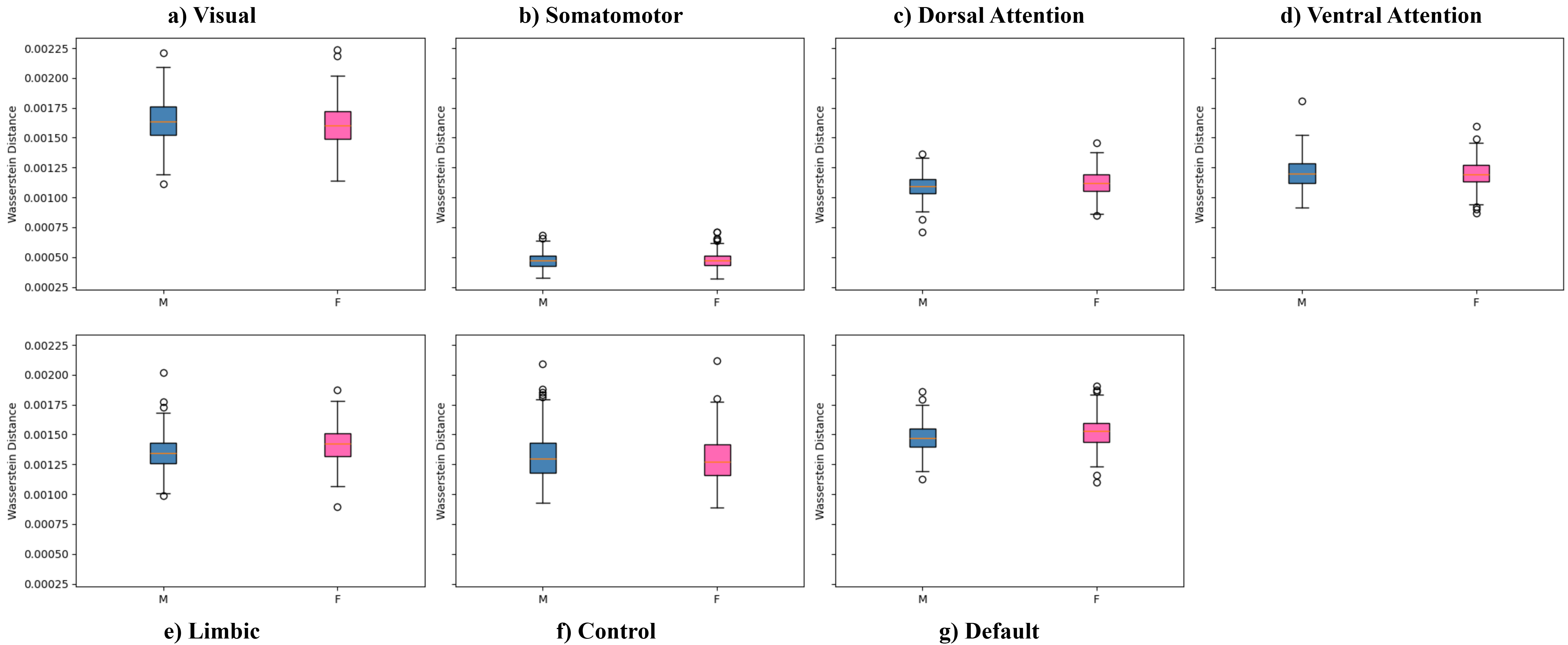}
\caption{Wasserstein distance between network cut off and baseline predictions: SC (network cut off) to predicted FC. X-Axis: Genders; Y-Axis: Wasserstein Distance (0.00025--0.00225). Blue: male, pink: female. Each subfigure shows the per-subject distribution for one Yeo network.}
\label{fig:wasssc2fc}
\end{figure}

In the SC (network cut off) $\rightarrow$ predicted FC direction, all networks show low eigenspectrum divergence. Visual and Somatomotor networks being zeroed out, produce the smallest spectral perturbations. Male and female KL distributions overlap almost completely across all networks. This is summarised in figures~\ref{fig:klsc2fc},~\ref{fig:frobsc2fc} and~\ref{fig:wasssc2fc}.

\begin{figure}[H]
\centering
\includegraphics[width=1\linewidth]{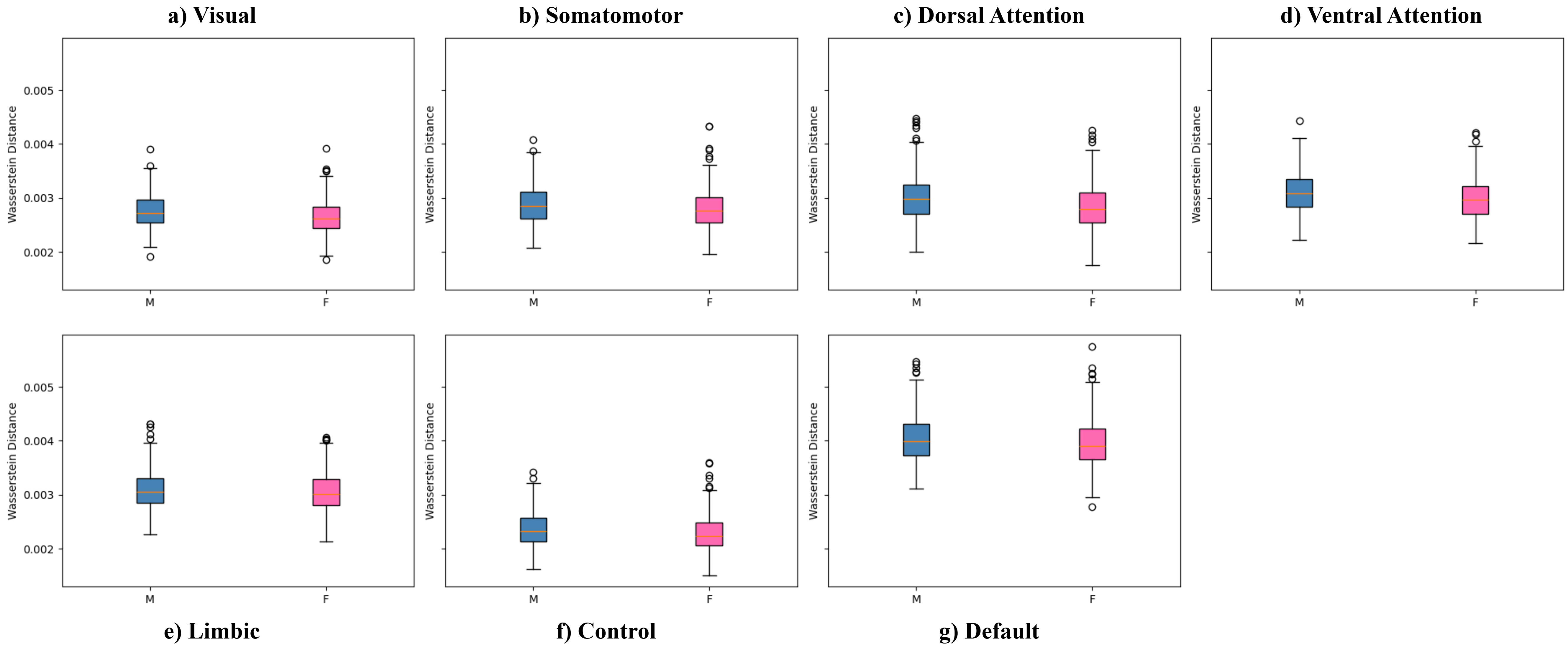}
\caption{Wasserstein distance between network cut off and baseline predictions: FC (network cut off) to predicted SC. X-Axis: Genders; Y-Axis: Wasserstein Distance (0.002--0.005). Blue: male, pink: female. Each subfigure shows the per-subject distribution for one Yeo network.}
\label{fig:wassfc2sc}
\end{figure}

In the FC (network cut off) $\rightarrow$ predicted SC direction, the Default Mode and Dorsal Attention Networks produce the largest spectral perturbations, with higher-order association networks showing more subject-level outliers than sensory networks. Females show slightly greater spread in DMN and Dorsal Attention outliers, though no clear sex separation is present at the population level. This is summarised in figure~\ref{fig:wassfc2sc}.

Across all three metrics and both prediction directions, the Default Mode Network consistently produced the largest perturbation when cut off. The Somatomotor network produced the smallest.

A consistent asymmetry exists between the two prediction directions: FC (network cut off) $\rightarrow$ predicted SC perturbations are substantially larger in absolute magnitude than SC (network cut off) $\rightarrow$ predicted FC perturbations, reflecting the known difference in prediction difficulty between the two directions.

\subsection{Sex Classification}

\textbf{N/w cut off features} For the FC-only feature set, Logistic Regression achieves the highest accuracy at $0.6480 \pm 0.0536$. For SC-only features, RandomForest leads at $0.5898 \pm 0.0427$. For combined SC+FC features, SVM RBF achieves $0.6609 \pm 0.0531$. The consistent improvement of combined features over either modality alone suggests complementary sex-discriminative information across directions.

\textbf{N/w intact baseline} Using the full predicted FC directly, LDA achieves $0.8362 \pm 0.0261$. Predicted SC alone peaks at $0.6068 \pm 0.0217$ with SVM RBF. The combined predicted FC+SC set yields $0.8476 \pm 0.0309$ with MLP. The substantially higher accuracy of these features confirms that the full predicted connectome retains considerably more sex-discriminative information than network-level perturbation signatures alone, and that the Krakencoder's predictions do not dilute this information.

The accuracies of various classifiers in classifying the two sexes with features obtained from predicted connectomes with their networks intact, and with the networks cut off, are summarised in tables~\ref{tab:allmodalities} and ~\ref{tab:results}.

\begin{table}[ht]
\centering
\caption{Classification Accuracy across Modalities}\label{tab:allmodalities}
\setlength{\tabcolsep}{4pt}
\begin{tabularx}{\textwidth}{l|XX|XX|XX}
\hline
\textbf{Classifier} & \multicolumn{2}{c|}{FC Only} & \multicolumn{2}{c|}{SC Only} & \multicolumn{2}{c}{SC + FC} \\
 & N/ws Cut off & N/ws Intact & N/ws Cut off & N/ws Intact & N/ws Cut off & N/ws Intact \\
\hline
Logistic Regression  & $\textbf{0.6480}$ & $0.8290$ & $0.5755$ & $0.5926$ & $0.6453$ & $0.8034$ \\
SVM RBF              & $0.6111$ & $0.8247$ & $0.5627$ & $0.6068$ & $\textbf{0.6609}$ & $0.8376$ \\
Random Forest        & $0.6039$ & $0.7763$ & $\textbf{0.5898}$ & $0.5968$ & $0.6481$ & $0.7850$ \\
KNN                  & $0.5755$ & $0.6680$ & $0.5312$ & $0.4958$ & $0.5955$ & $0.6524$ \\
AdaBoost             & $0.5854$ & $0.8020$ & $0.5456$ & $0.5470$ & $0.6424$ & $0.7806$ \\
Gradient Boosting    & $0.6069$ & $0.7906$ & $0.5599$ & $0.5670$ & $0.6410$ & $0.7991$ \\
LDA                  & $0.6352$ & $\textbf{0.8362}$ & $0.5698$ & $\textbf{0.5982}$ & $0.6410$ & $0.7991$ \\
Naive Bayes          & $0.6025$ & $0.7165$ & $0.5756$ & $0.5868$ & $0.6224$ & $0.7236$ \\
MLP                  & $0.5969$ & $0.8290$ & $0.5327$ & $0.5799$ & $0.6053$ & $\textbf{0.8476}$ \\
XGBoost              & $0.6125$ & $0.7919$ & $0.5556$ & $0.5769$ & $0.6225$ & $0.8162$ \\
\hline
\end{tabularx}
\end{table}

\begin{table}[ht]
\centering
\caption{Summary of sex classification results obtained}\label{tab:results}
\setlength{\tabcolsep}{12pt}
\makebox[\textwidth][c]{%
\begin{tabular}{l|l|l}
\toprule
\textbf{Condition} & \textbf{Feature Set} & \textbf{Best Accuracy (Classifier)} \\
\midrule
\multirow{3}{*}{Networks Cut off}
  & FC Only        & $0.6480 \pm 0.054$ (Logistic Regression) \\
  & SC Only        & $0.5898 \pm 0.042$ (Random Forest) \\
  & SC+FC Combined & $0.6609 \pm 0.053$ (SVM RBF) \\
\midrule
\multirow{3}{*}{Networks Intact}
  & FC only        & $0.8362 \pm 0.026$ (LDA) \\
  & SC only        & $0.6068 \pm 0.022$ (SVM RBF) \\
  & SC+FC Combined & $0.8476 \pm 0.031$ (MLP) \\
\bottomrule
\end{tabular}
}
\end{table}

\section{Discussion}

The Default Mode Network's consistently large perturbation reflects its well-characterised structural-functional correspondence and its central role in the Krakencoder's cross-modal mapping. Removing its functional connectivity causes the largest redistribution of predicted structural connectivity, consistent with the Krakencoder's own internal sensitivity analysis. The Somatomotor network's low perturbation reflects its reliance on indirect polysynaptic pathways, meaning zeroing its functional connectivity has a smaller impact on structural prediction.

Sex differences in network-level perturbation signatures were present but subtle. Male and female distributions were closely matched across all networks, metrics, and prediction directions, with only marginal differences in higher-order networks such as the DMN and Dorsal Attention Network. These subtle differences were insufficient to produce strong sex classification from network cut off features alone. The strong performance of predicted FC obtained from SCs with their networks intact relative to predicted SC is consistent with the broader literature showing that functional connectivity is more directly sex-discriminative than structural connectivity.

\subsection{Limitations}
Several limitations should be noted. The analysis is constrained by the choice of the FS86 parcellation, and results may differ under alternative atlases with different spatial resolutions or region definitions. More broadly, the granularity of all findings is bounded by the number of regions considered: It is only as course as the number of ROIs. Additionally, the Yeo-7 network definitions used for cutting off networks assume non-overlapping network membership, and do not account for regions that may participate in multiple functional networks simultaneously, which is a known property of cortical organisation. Finally, the Krakencoder was trained on healthy young adults, and the perturbation signatures reported here reflect the model's learned mapping in that population, which may not generalise to clinical or lifespan cohorts without retraining or fine-tuning.

\section{Conclusion and Future Work}

This study applied a network zeroing approach to structural and functional connectomes using the Krakencoder to investigate sex-based differences in brain network organisation. The Default Mode Network produced the largest perturbation across all metrics and both prediction directions. Sex differences in perturbation signatures were present but subtle, with features obtained from predicted connectomes whose inputs had a network cut off achieving at most 66.09\% classification accuracy versus 84.76\% from unperturbed predicted connectomes. These findings validate the network-cut-off approach as a tool for exploring structure-function relationships.

Future work could extend this analysis to clinical populations, apply region-level rather than network-level cutting-off, and explore multiple connectome flavours to establish robustness across methodological choices.

\printbibliography

\section*{Supplementary}\label{sec:supplementary}

There are distinct ways of constructing a brain connectivity matrix, reflecting methodological choices in data processing. In this study, we use the term \textit{connectome flavour} to denote a specific combination of 
(1) parcellation scheme, 
(2) functional connectivity (FC) estimation method, and 
(3) structural connectivity (SC) estimation method. 
Each flavour represents one of these methodological choices. The Krakencoder was trained to translate between these flavours via a shared latent representation.

\subsection*{Parcellation Schemes}
\begin{itemize}
    \item \textbf{FS86 (FreeSurfer 86 atlas):} Derived from the Desikan--Killiany atlas, consisting of 68 cortical regions and 18 subcortical/cerebellar regions. Provides anatomically interpretable parcellation at moderate resolution.
    \item \textbf{Shen268 atlas:} A functional parcellation into 268 regions obtained by clustering resting-state fMRI data. Offers finer granularity for functional network analysis.
    \item \textbf{Coco439 atlas:} A high-resolution parcellation with 439 regions. Captures subtle connectivity patterns and inter-individual variability, though at higher computational cost.
\end{itemize}

\subsection*{Functional Connectivity Estimates}
\begin{itemize}
    \item \textbf{Pearson correlation:} Standard measure of linear association between regional time series. Produces dense FC matrices but is sensitive to global confounds.
    \item \textbf{Global signal regression (GSR):} Correlation computed after regressing out the global mean time series. Reduces widespread noise and motion artifacts, though it alters the distribution of FC values.
    \item \textbf{Partial correlation:} Estimates direct statistical dependencies between regions while controlling for all others. Produces sparser FC matrices and reduces indirect effects.
\end{itemize}

\subsection*{Structural Connectivity Estimates}
\begin{itemize}
    \item \textbf{Deterministic tractography:} Streamline reconstruction following the principal diffusion direction voxel by voxel. Provides reproducible pathways but limited in resolving crossing fibers.
    \item \textbf{Probabilistic tractography:} Samples multiple possible streamline paths based on local fiber orientation distributions. Captures uncertainty and crossing fibers more effectively, producing richer but noisier connectivity estimates.
\end{itemize}

\begin{figure}[ht]
\centering
\includegraphics[width=1\linewidth]{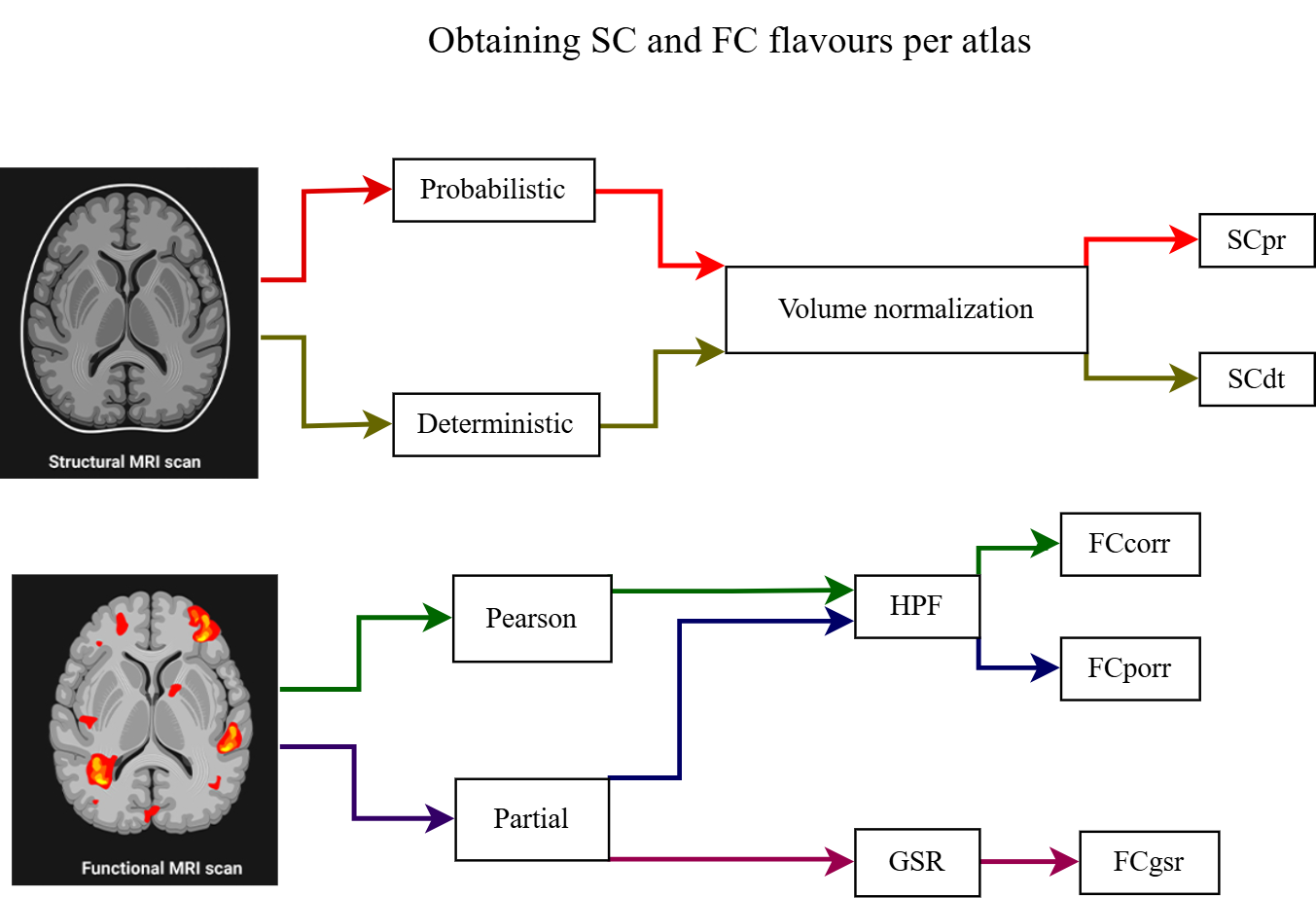}
\caption{Obtaining different flavours of connectomes from MRI data}
\label{fig:flavs}
\end{figure}

The figure shows us different pipelines used to obtain the 15 flavours that correspond to one encoder and one decoder each. This gives us 15 encoders and decoders in total. Each colour represents one pipeline. In order,

\begin{itemize}
    \item Structural MRI $\rightarrow$ Probabilistic tractography $\rightarrow$ Volume normalization $\rightarrow$ SCpr
    \item Structural MRI $\rightarrow$ Deterministic tractography $\rightarrow$ Volume normalization $\rightarrow$ SCdt
    \item Functional MRI $\rightarrow$ Pearson correlation $\rightarrow$ High-pass filtering (HPF) $\rightarrow$ FCcorr
    \item Functional MRI $\rightarrow$ Partial correlation $\rightarrow$ High-pass filtering (HPF) $\rightarrow$ FCporr
    \item Functional MRI $\rightarrow$ Partial correlation $\rightarrow$ Global signal regression (GSR) $\rightarrow$ FCgsr
\end{itemize}

Note: the violet arrow in the flowchart indicates that the Pearson correlation and Partial correlation pipelines share a common initial step (Partial correlation) before diverging into separate estimation methods.

The detailed sex classification results (classification of sexes with 5-fold cross validation) are shown in tables ~\ref{tab:fconly}, ~\ref{tab:sconly} and ~\ref{tab:scplusfc}.

\begin{table}[ht]
\centering
\caption{Classification Accuracy of Male and Female Classes: FC Only}\label{tab:fconly}
\setlength{\tabcolsep}{12pt}
\begin{tabular}{l|c|c}
\hline
\textbf{Classifier} & \textbf{N/ws cut off} & \textbf{N/ws intact} \\
\hline
Logistic Regression  & $0.6480 \pm 0.0536$ & $0.8290 \pm 0.0329$ \\
SVM RBF              & $0.6111 \pm 0.0292$ & $0.8247 \pm 0.0232$ \\
Random Forest        & $0.6039 \pm 0.0281$ & $0.7763 \pm 0.0150$ \\
KNN                  & $0.5755 \pm 0.0210$ & $0.6680 \pm 0.0257$ \\
AdaBoost             & $0.5854 \pm 0.0455$ & $0.8020 \pm 0.0124$ \\
Gradient Boosting    & $0.6069 \pm 0.0438$ & $0.7906 \pm 0.0300$ \\
LDA                  & $0.6352 \pm 0.0639$ & $0.8362 \pm 0.0261$ \\
Naive Bayes          & $0.6025 \pm 0.0391$ & $0.7165 \pm 0.0286$ \\
MLP                  & $0.5969 \pm 0.0098$ & $0.8290 \pm 0.0277$ \\
XGBoost              & $0.6125 \pm 0.0380$ & $0.7919 \pm 0.0383$ \\
\hline
\end{tabular}
\end{table}

\begin{table}[ht]
\centering
\caption{Classification Accuracy of Male and Female Classes: SC Only}\label{tab:sconly}
\setlength{\tabcolsep}{12pt}
\begin{tabular}{l|c|c}
\hline
\textbf{Classifier} & \textbf{N/ws cut off} & \textbf{N/ws intact} \\
\hline
Logistic Regression  & $0.5755 \pm 0.0330$ & $0.5926 \pm 0.0228$ \\
SVM RBF              & $0.5627 \pm 0.0313$ & $0.6068 \pm 0.0217$ \\
Random Forest        & $0.5898 \pm 0.0427$ & $0.5968 \pm 0.0213$ \\
KNN                  & $0.5312 \pm 0.0570$ & $0.4958 \pm 0.0331$ \\
AdaBoost             & $0.5456 \pm 0.0457$ & $0.5470 \pm 0.0308$ \\
Gradient Boosting    & $0.5599 \pm 0.0408$ & $0.5670 \pm 0.0351$ \\
LDA                  & $0.5698 \pm 0.0323$ & $0.5982 \pm 0.0354$ \\
Naive Bayes          & $0.5756 \pm 0.0351$ & $0.5868 \pm 0.0336$ \\
MLP                  & $0.5327 \pm 0.0381$ & $0.5799 \pm 0.0378$ \\
XGBoost              & $0.5556 \pm 0.0157$ & $0.5769 \pm 0.0554$ \\
\hline
\end{tabular}
\end{table}

\begin{table}[ht]
\centering
\caption{Classification Accuracy of Male and Female Classes: SC + FC Combined}\label{tab:scplusfc}
\setlength{\tabcolsep}{12pt}
\begin{tabular}{l|c|c}
\hline
\textbf{Classifier} & \textbf{N/ws cut off} & \textbf{N/ws intact} \\
\hline
Logistic Regression  & $0.6453 \pm 0.0348$ & $0.8034 \pm 0.0291$ \\
SVM RBF              & $0.6609 \pm 0.0531$ & $0.8376 \pm 0.0200$ \\
Random Forest        & $0.6481 \pm 0.0359$ & $0.7850 \pm 0.0483$ \\
KNN                  & $0.5955 \pm 0.0305$ & $0.6524 \pm 0.0176$ \\
AdaBoost             & $0.6424 \pm 0.0198$ & $0.7806 \pm 0.0395$ \\
Gradient Boosting    & $0.6410 \pm 0.0184$ & $0.7991 \pm 0.0242$ \\
LDA                  & $0.6410 \pm 0.0458$ & $0.7991 \pm 0.0455$ \\
Naive Bayes          & $0.6224 \pm 0.0571$ & $0.7236 \pm 0.0318$ \\
MLP                  & $0.6053 \pm 0.0225$ & $0.8476 \pm 0.0309$ \\
XGBoost              & $0.6225 \pm 0.0398$ & $0.8162 \pm 0.0348$ \\
\hline
\end{tabular}
\end{table}

\end{document}